\documentclass[runningheads]{template/llncs}
\usepackage{template/eccv}

\usepackage{template/eccvabbrv}

\usepackage{graphicx}
\usepackage{booktabs}
\usepackage[accsupp]{axessibility} 

\usepackage[pagebackref,breaklinks,colorlinks,citecolor=eccvblue]{hyperref}
\usepackage{orcidlink}

\newcommand{\argmin}{\mathop{\rm argmin}}

\newcommand{\bc}{{\mathbf c}}
\newcommand{\ct}{{\mathbf c}_{\theta}}

\newcommand{\black}{\color{black}}

\usepackage{amsmath, amssymb, amsfonts}
\usepackage{graphicx}
\usepackage{booktabs}
\usepackage[dvipsnames]{xcolor}
\usepackage{bm}
\usepackage{algorithm, algpseudocode}
\usepackage{colortbl}
\usepackage{bbm}
\usepackage{makecell}
\usepackage{threeparttable}
\usepackage{multirow}
\usepackage{wrapfig}

\newcommand{\cond}{\bm{h}}

\begin{document}

\title{Bridging Vision and Language Concepts through \\ Optimal Transport Semantic Flow} 

\author{
Chenyang Zhang\inst{1}$^{*}$ \and
Anqi Dong\inst{2}$^{*}$ \and
Guangming Zhu\inst{1} \and
Nuoye Xiong\inst{1} \and
Siyuan Wang\inst{1} \and
Lin Mei\inst{1} \and
Liang Zhang\inst{1}$^{\dagger}$
}

\authorrunning{Zhang et al.}

\institute{
School of Computer Science and Technology, \\
Xidian University, Xi'an, China
\and
KTH Royal Institute of Technology, \\
Stockholm, Sweden \\
\email{chenyang.zhang@stu.xidian.edu.cn, anqid@kth.se, liangzhang@xidian.edu.cn}
}

\maketitle

\renewcommand{\thefootnote}{\fnsymbol{footnote}}
\footnotetext[0]{$^{*}$ Equal Contribution.\quad $\dagger$ Corresponding author.}

\begin{abstract}
Concept Bottleneck Models (CBMs) promise transparent reasoning by predicting through human-interpretable concepts, yet their effectiveness fundamentally depends on how well visual and textual representations are aligned or matched. Existing vision–language CBMs often rely on pre-aligned encoders or global cosine similarity, which obscures fine-grained concept localization and fails to reflect true semantic geometry.  In this work, we rethink concept alignment as dynamic cross-modal transport process instead of static projection and propose Optimal Transport Flow Concept Bottleneck Model \textbf{(OTF-CBM)}. It first learns a data-driven semantic cost via Inverse Optimal Transport to measure cross-modal distances, and then performs unbalanced optimal-transport-based flow matching to model semantic transitions between visual patches and textual concepts.  With velocity-based concept activation, OTF-CBM captures interpretable geometric relations without ODE integration. Experiments further show that OTF-CBM achieves great classification accuracy and concept faithfulness, offering a new geometric and dynamical perspective for interpretable cross-modal reasoning. Our code will be released at \href{https://github.com/ChenyangZhang00/OTF-CBM/}{https://github.com/ChenyangZhang00/OTF-CBM}.

\end{abstract}

\begin{figure}[t]
\centering
\includegraphics[width=1\linewidth]{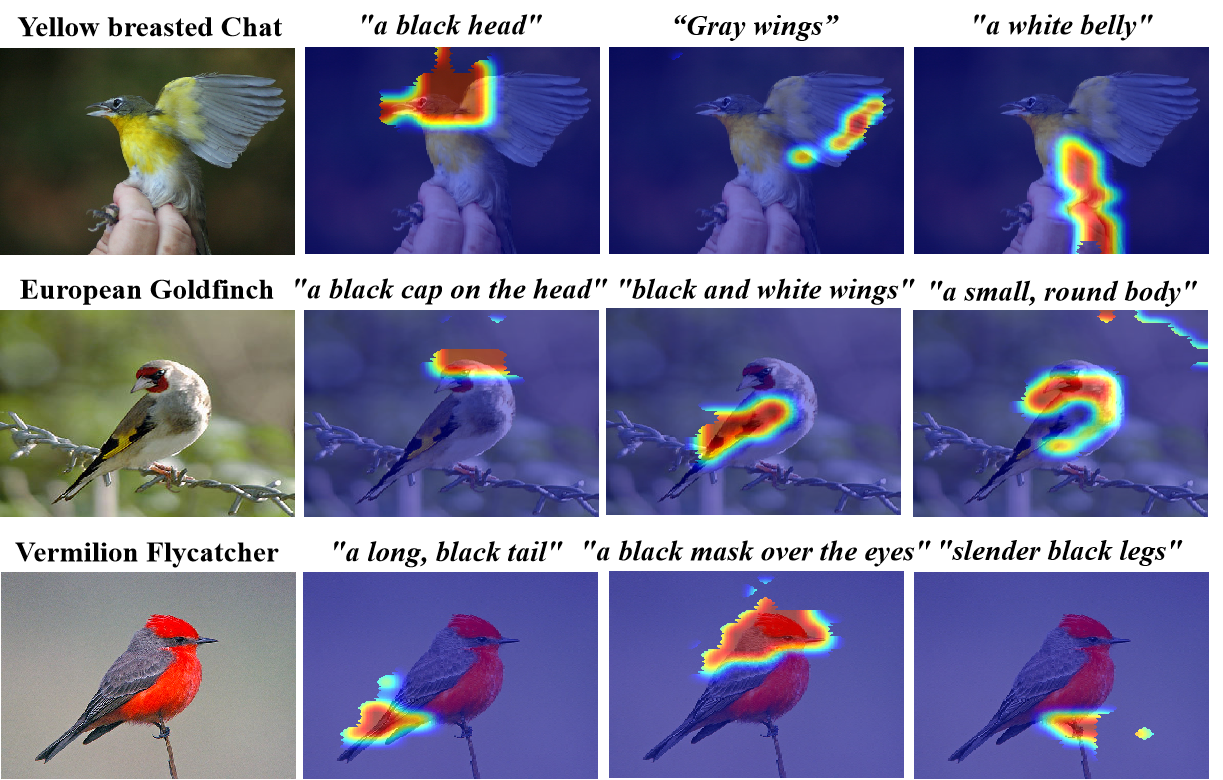}
\caption{
\textbf{Cross-modal concept visualization with OTF-CBM.} The model localizes fine-grained parts (head, wings, legs) and aligns them with textual concepts. Compared with prior CBMs, it yields more coherent, spatially grounded components and smooth semantic flow from visual features to concept embeddings.
}
\label{fig:4}
\vspace{-1.8em}
\end{figure} 

\vspace{-1cm}
\section{Introduction}\label{sec:intro}
Concept bottleneck models expose an intermediate layer of human concepts --- a model first predicts concept activations and then uses them to make final decisions~\cite{koh2020concept, yuksekgonul2022post, vandenhirtz2024stochastic}. This structure enables inspection, diagnosis, and targeted intervention. In vision--language settings, however, three limitations persist:
(i) \emph{Fixed embedding dependence:} reusing a shared CLIP space ties the visual encoder to a linguistic geometry that need not match concept granularity~\cite{radford2021learning, yang2023language, srivastava2024vlg}, (ii) \emph{Global--to--concept inference:} predicting all concept scores from a single global feature blurs spatial evidence and weakens localization~\cite{oikarinen2023label,yang2023language,srivastava2024vlg}, and (iii) \emph{Inadequate similarity:} cosine similarity aligns category labels trained by large contrastive objectives but is a poor proxy for fine region--to--concept relations~\cite{radford2021learning}.

\begin{figure}
\centering
\includegraphics[width=0.8\linewidth]{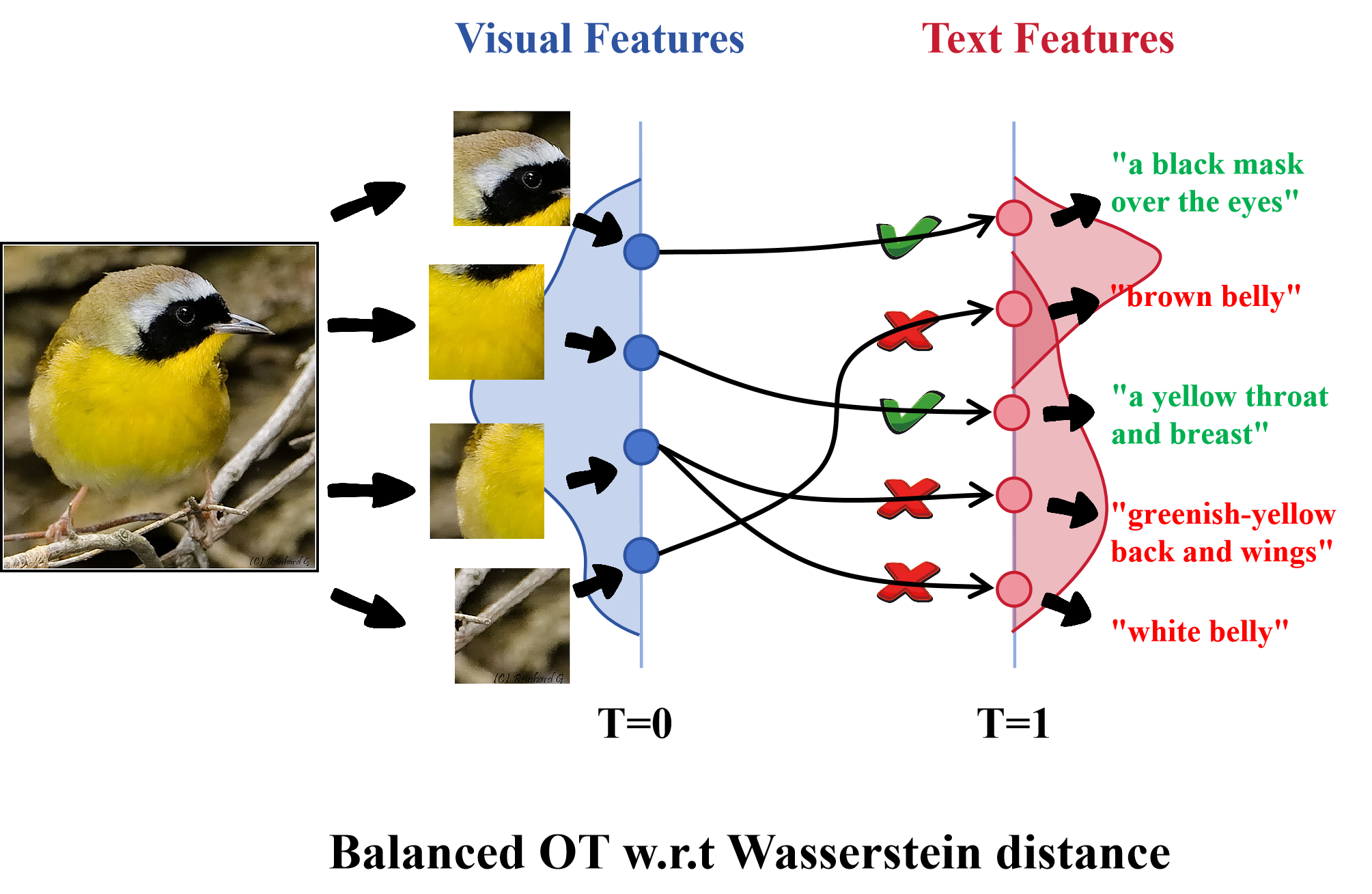}
\caption{\textbf{Failure of standard OT for cross-modal concept matching.}
With fixed ground cost, standard OT often misaligns visual patches and concepts in heterogeneous spaces.}
\label{fig:1}
\end{figure}

Effective CBM models should move beyond static projection and be able to construct a flexible, learnable cross-modal geometry that explicitly connects visual regions with textual concepts. Such a formulation requires establishing region-to-concept correspondences that are both semantically accurate and geometrically consistent across modalities. This introduces two key challenges: (i) discovering fine-grained correspondences between image patches and textual concepts in the absence of explicit component-level annotations, and (ii) modeling how these correspondences evolve into interpretable conceptual relations within a unified visual--textual space. A model capable of capturing both the spatial grounding of concepts and the dynamics of their semantic transitions would enable a more faithful and transparent conceptual reasoning process.

\begin{figure*}[t]
  \centering
  \includegraphics[width=\linewidth]{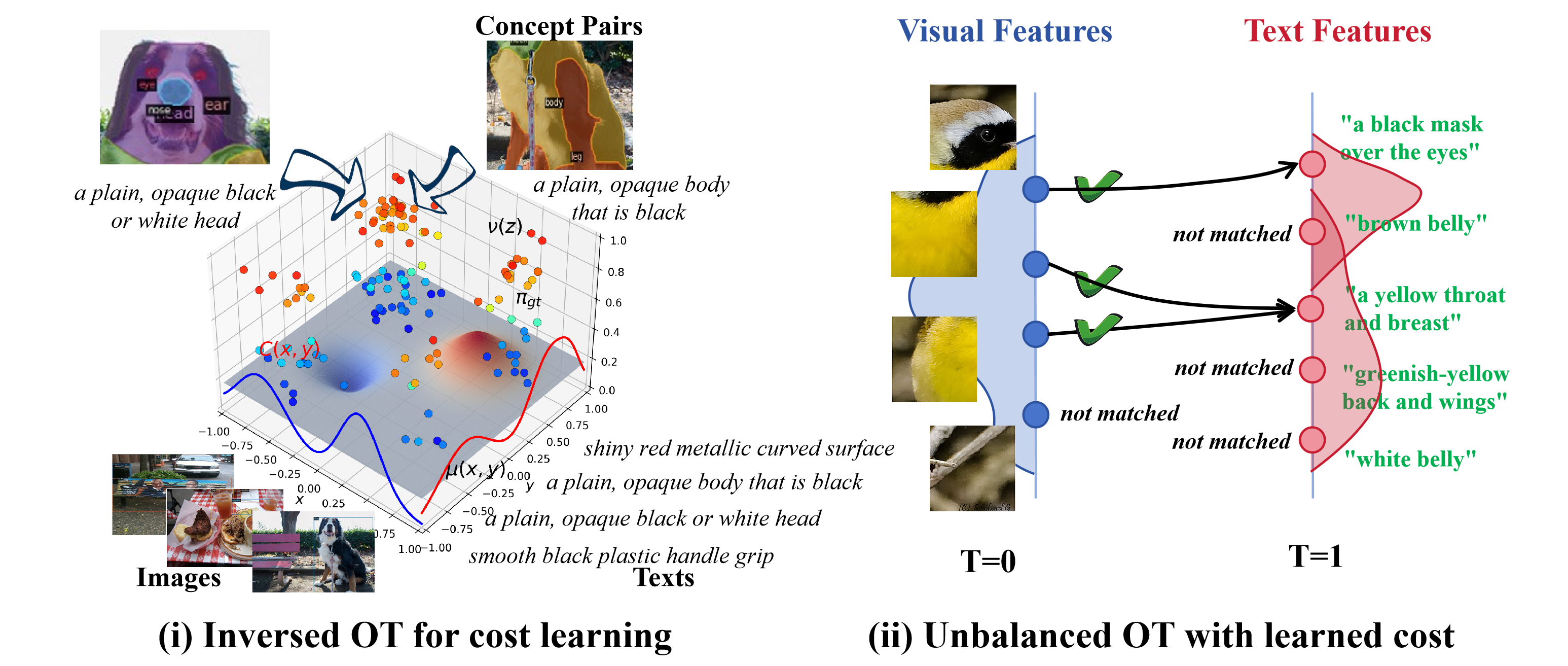}
  \caption{
\textbf{Cross-modal concept matching with our solution.}
(i) We address this by learning a data-driven cost $\bc_\theta$ from component annotations, yielding semantically calibrated metrics. 
(ii) Combining learned cost with unbalanced OT enables flexible many-to-one and partial matchings suited to vision–language alignment.
}
  \label{fig:1_1}
\vspace{-0.2in}
\end{figure*} 

To this end, we propose OTF-CBM, a variant of CBM that explicitly models cross-modal geometry (Fig.~\ref{fig:4}). We first cast region--to--concept alignment as an optimal transport (OT) problem~\cite{villani2009optimal,peyre2019computational} between unaligned visual and textual encoders, determining how visual evidence should be assigned to semantic concepts. Next, OTF-CBM learns a data-driven cross-modal cost via Inverse Optimal Transport (IoT)~\cite{li2019learning, ma2020learning,chiu2022discrete} from annotated region--concept associations, rather than relying on hand-crafted distances such as cosine or Euclidean metrics (Fig.~\ref{fig:1}). This yields a cost landscape under which the optimal transport plan reproduces true semantic correspondences. Since visual--textual mappings are inherently unbalanced (e.g., multiple patches may correspond to a single concept, and many background regions to none), we adopt an Unbalanced OT (UOT) formulation \cite{chizat2016scaling,sejourne2023unbalanced,de2023unbalanced}, allowing partial and many-to-one matching to emerge naturally. As shown in Fig.~\ref{fig:1_1}, the resulting transport geometry thus provides a faithful semantic metric between modalities and forms the basis of our concept reasoning.

Building upon on this learned geometry, we train a conditional velocity field through OT-based Flow Matching (FM) \cite{lipman2022flow}. Instead of enforcing static similarity in a shared embedding space, the velocity field models the \emph{dynamic semantic flow} from visual prototypes to textual concepts.\footnote{Throughout, we use \emph{visual, region prototypes}, and \emph{clusters} interchangeably}
At inference time, OTF-CBM does not integrate an ODE to reconstruct a final state. Instead, it introduces a velocity-based concept activation mechanism that compares the predicted velocity at an intermediate time step with the ideal velocity toward each concept. This single-step, geometry-aware activation replaces conventional cosine similarity, capturing how visual evidence \emph{moves} toward conceptual meaning and yielding concept predictions that are both more accurate and more interpretable than prior CBM variants \cite{koh2020concept,yuksekgonul2022post,vandenhirtz2024stochastic,radford2021learning}. The contributions are as follows: 
\begin{enumerate}
\item[i)] \emph{Cross-modal optimal transport for semantic correspondence.} 
We introduce an Optimal Transport framework that establishes fine-grained semantic correspondences between visual and textual modalities, integrating Inverse OT for adaptive cost learning and Unbalanced OT for non-uniform mass transfer.

\item[ii)] \emph{Flow-based perspective on concept alignment.} 
We reformulate concept reasoning as a dynamic transformation process driven by a learned velocity field, with geometry-aware \emph{semantic flow} that infers concept activation from directional consistency.

\item[iii)] \emph{Great empirical performance.} 
Experiments show that OTF-CBM outperforms state-of-the-art CBMs in accuracy, interpretability, and generality.
\end{enumerate}\black

\section{Related work}\label{sec:relate}

Concept-based interpretability aims to make predictions intervenable by routing decisions through human-understandable variables. Classical CBMs~\cite{koh2020concept} factor prediction into concept extraction and label prediction so that downstream decisions depend on concepts. Early CBMs rely on expert-annotated attributes, which limit scalability and transfer. With large vision–language models such as CLIP~\cite{radford2021learning} and GPT-style LLMs, recent works automate concept construction and supervision. Post-hoc CBM (P-CBM)~\cite{yuksekgonul2022post} projects visual embeddings onto a CLIP-derived concept bank. Label-Free CBM~\cite{oikarinen2023label} uses CLIP pseudo labels to train concept predictors without human annotation. LaBo~\cite{yang2023language} employs LLMs to propose diverse candidate concepts and applies submodular selection to refine the bank. PCBM-h~\cite{shang2024incremental} adds a residual head to refine concept predictions and improves accuracy at the expense of interpretability. These approaches improve scalability and performance but still impose global features on concept mappings that mix localized evidence and hinder fine-grained grounding. DOT\mbox{-}CBM casts CBM learning as optimal transport between image patches and concept embeddings, yielding spatially grounded concepts~\cite{xie2025discovering}. Our approach differs by learning  cross\mbox{-}modal cost and using unbalanced OT, which accommodates heterogeneous semantics and many\mbox{-}to\mbox{-}one or background correspondences.

Optimal Transport theory~\cite{peyre2019computational, villani2021topics,chen2021stochastic} matches source distribution $\mu(x)$ and target distribution $\nu(y)$ by minimizing transport cost $\bc(x,y)$ over the set of couplings $\Pi(\mu,\nu)\;=\;\bigl\{\pi\ge 0 \; |  \pi\mathbf{1}=\mu,\ \mathbf{1}^\top\pi=\nu \bigr\}$.
The Kantorovich formulation \cite{rachev1998mass,dong2024monge} reads
$$
\min_{\pi\in\Pi(\mu,\nu)}\ \langle \bc,\pi\rangle
\;\equiv\;\iint_{\mathcal{X}\times\mathcal{Y}} \bc(x,y)\, \mathrm{d}\pi(x,y),
$$
and its entropic regularization~\cite{cuturi2013sinkhorn,dong2025data,dong2024network,dong2025negative} with regularizer $\varepsilon \pi \log\pi$ for computational efficiency, solving scalable Sinkhorn iterations and widely used in vision and multi-modal learning~\cite{courty2018joint}. When masses are imbalanced, unbalanced OT (UOT) \cite{chizat2018scaling,sejourne2023unbalanced,chizat2018unbalanced,fatras2021unbalanced} relaxes the marginal constraints by penalizing deviations with divergences, enabling robust alignment in heterogeneous settings. Inverse Optimal Transport (IOT)~\cite{galichon2016optimal,li2019learning} seeks to \emph{learn} a cost function that explains observed matching. Given an empirical coupling $\hat\pi$ (from paired data) and a parameterized cost $\bc_\theta$, a practical formulation fits $\ct$ by matching the entropic OT plan $\pi_\theta$ induced by $\bc_\theta$ to $\hat\pi$ that reads 
$$
\min_{\theta} \mathcal{D}\!\left(\hat\pi,\pi_\theta\right) + \lambda\,\mathcal{R}(\theta) \quad \mbox{s.t.} \quad \pi_\theta\in\argmin_{\pi\in\Pi(\mu,\nu)} \langle \bc_\theta,\pi\rangle+\varepsilon\,\pi\log\pi,
$$
where $\mathcal{D}$ is a discrepancy (e.g., KL or $\ell_2$ between couplings), and $\mathcal{R}$ regularizes the cost class for stability and identifiability up to admissible gauges. Extensions include contrastive metric learning and multi-modal representation learning that learn semantically structured costs.

Flow Matching (FM)~\cite{lipman2022flow,albergo2022building,dong2026meanflow} learns a time-dependent velocity field that transports a sample $x_0$ from a source distribution toward a target sample $x_1$ without stochastic simulation. A common choice uses straight-line paths $x_t=(1-t)\,x_0+t\,x_1$ for $t\in[0,1]$ and the target velocity $u_t=x_1-x_0$. A parametric field $v_\phi(x,t,\mathrm{cond})$ is trained by the regression loss
$$
\mathcal{L}_{\mathrm{FM}}
=\mathbb{E}_{(x_0,x_1)\sim\Pi,\ t\sim\mathcal{U}[0,1]}
\Bigl[\|\,v_\phi(x_t,t,\mathrm{cond})-u_t\,\|^2\Bigr],
$$
where $\pi$ specifies how training pairs are drawn and $\mathrm{cond}$ denotes optional conditioning. OT-consistent variants choose $\Pi$ via minibatch OT or UOT so the supervised trajectories reflect transport geometry. OT-CFM~\cite{tong2023improving,yue2025oat} integrates minibatch OT couplings to accelerate convergence and promote OT-consistent dynamics, while CrossFlow~\cite{liu2025flowing} extends FM to cross-modal generation without explicit noise injection. We also note that most existing OT and FM pipelines assume fixed costs or balanced marginals.

\section{Method}\label{sec:method}

We start from a standard concept bottleneck model (CBM) and replace its similarity-based concept inference with a geometry-aware matching mechanism. 

In Sec.~\ref{subsec:overview}, we set up the CBM pipeline and highlight the limitation of pointwise similarity for concept prediction. The goal is simple: given image features and concept embeddings, we want a reliable way to (i) align the two modalities, and (ii) turn that alignment into stable concept activations.

Section~\ref{sec:vlot} introduces Visual–Language Optimal Transport (VLOT), which builds an explicit cross-modal coupling between visual and concept embeddings. VLOT uses an unbalanced transport formulation and a background-aware penalty so the coupling can ignore irrelevant image content and handle missing or extra mass across modalities.

Section~\ref{subsec:iot} makes the coupling meaningful by learning the transport cost from data. We do this via inverse optimal transport, so the resulting transport plans reflect the geometry that is actually useful for semantic matching, rather than relying on a hand-designed distance.

Finally, Section~\ref{subsec:3.4} turns the discrete transport solution into a continuous concept-inference rule. We train a straight-line velocity field using transport displacements as supervision, and we compute concept activations from how well the predicted instantaneous velocity agrees with the transport direction. This produces smooth, robust concept scores that plug directly.

\subsection{Problem setting and notation}\label{subsec:overview}

Classical visual classifiers learn a direct mapping 
$f:\mathcal{X}\rightarrow\mathcal{Y}$ 
to approximate $P(y|x)$ from data 
$\{(x_i,y_i)\}_{i=1}^N$, 
but provide limited interpretability. Concept Bottleneck Models (CBMs) introduce an intermediate concept representation.  Given a predefined concept set 
$\mathcal{C}=\{c_j\}_{j=1}^M$, 
CBMs learn a two-stage mapping
$f_{X\rightarrow C}:\mathcal{X}\rightarrow\mathbb{R}^M$ and 
$f_{C\rightarrow Y}:\mathbb{R}^M\rightarrow\mathcal{Y}$.
An input image is first mapped to a concept activation vector 
$\mathbf{a}\in\mathbb{R}^M$, which is then used for classification.
\begin{table}
\centering
\begingroup
\setlength{\tabcolsep}{5pt}
\renewcommand{\arraystretch}{1.12}
\begin{tabular}{ll}
\toprule
Symbol & Meaning \\
\midrule
$\mathcal{X},\ \mathcal{C}$ & feature, concept space \\
$\mu,\ \nu$ & empirical distributions on $\mathcal{X}$ and $\mathcal{C}$ \\
$f_\psi$ & modality adapters to $\mathbb{R}^{d_p}$ \\
$\bc_\theta(x,c)$ & learned cross–modal transport cost \\
$\pi_\theta$ & OT or UOT plan induced by $\bc_\theta$ \\
$v_\phi(x,t\,|\,\cond)$ & conditional velocity field in FM \\
$\cond$ & conditioning input in $\mathbb{R}^{d_p}$ \\
$\mathbf{a}$ & concept activation vector \\
$f_{\mathrm{cls}},\ \hat y$ & classifier on concepts, predicted label \\
\bottomrule
\end{tabular}
\endgroup
\caption{Notation overview.}
\label{tab:notation-overview}
\end{table}
In vision--language CBMs, concept activations are typically 
computed via similarity in a shared embedding space, 
which may fail to capture structured cross-modal relations 
under complex distributions. Our method retains this bottleneck structure but replaces similarity-based concept inference with a geometry-aware transport and flow mechanism. The notations are summarized in Table \ref{tab:notation-overview}.

\begin{figure*}[t]
\centering
\includegraphics[width=\linewidth]{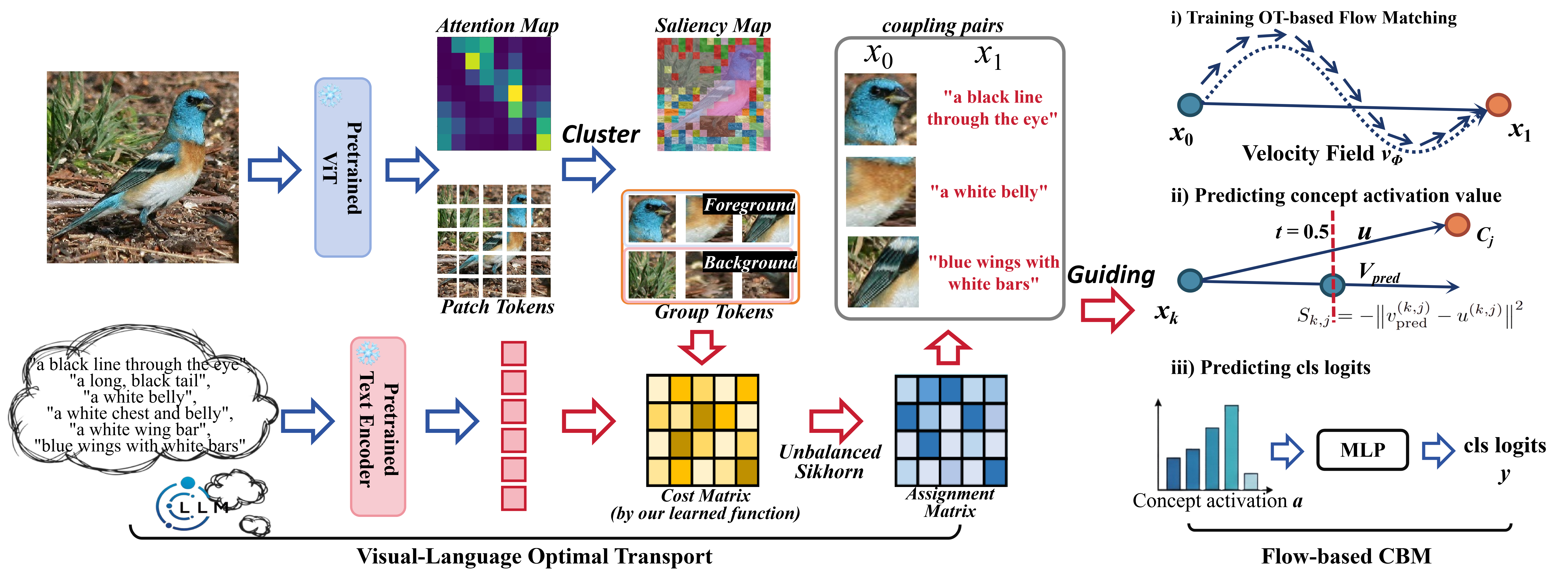}
\caption{ 
\textbf{Forward pipeline.}
Patch tokens are clustered into foreground and background. The learned cost $\bc_{\theta^\ast}$ forms a cost matrix to fixed concept embeddings with background penalties. Unbalanced OT yields a plan $\pi$. Samples from $\pi$ to train a conditional velocity field. At inference, concept activations come from midpoint velocity alignment, then a concept classifier produces labels.
}
\label{fig:3}
\end{figure*}

\subsection{Geometric Coupling via Vision--Language OT}
\label{sec:vlot}

To connect arbitrary visual and textual encoders, we cast cross-modal alignment as an optimal transport (OT) problem. Unlike cosine similarity, OT returns an explicit coupling that describes how visual evidence is distributed over semantic concepts under a learned geometry. Given an image, let $x_{1:N}=\{x_i\}_{i=1}^N\subset\mathcal{X}$ denote visual patch embeddings and let $c_{1:M}=\{c_j\}_{j=1}^M\subset\mathcal{C}$ denote textual concept embeddings. A transport plan $\pi\in\mathbb{R}_+^{N\times M}$ assigns nonnegative mass $\pi_{ij}$ from patch $x_i$ to concept $c_j$.

\noindent \textbf{Region aggregation.} Patch-level transport can be noisy because many neighboring patches describe the same region. We therefore cluster patch embeddings within each image using K-means and form $K$ group tokens by averaging features inside each cluster. Let $\tilde x_{1:K}=\{\tilde x_k\}_{k=1}^K$ denote these prototype embeddings. Transport is then computed between prototypes and concepts, which reduces redundancy and encourages region-level rather than patch-level alignment.

\noindent\textbf{Vision--Language OT (unbalanced).} A direct application of balanced OT enforces strict marginal conservation as in $\Pi(\mu,\nu)\;=\;\bigl\{\pi\ge 0 \; |  \pi\mathbf{1}=\mu,\ \mathbf{1}^\top\pi=\nu \bigr\}$,
which forces all visual mass to match some concept and forces all concepts to receive mass. This assumption is routinely violated in vision--language alignment: multiple regions may correspond to a single concept (many-to-one), background regions often have no semantic counterpart, and some concepts may be absent from the image. Under strict conservation, background prototypes are pushed into arbitrary matches, and absent concepts must absorb visual mass, leading to hallucinated correspondences.

We therefore adopt unbalanced optimal transport (UOT), which relaxes these marginal constraints via KL penalties. Let the prototype--concept cost matrix be
$(\bc_\theta)_{kj}=\bc_\theta\big(f_\psi(\tilde x_k),c_j\big)$. We compute the coupling as
\[
\pi_\theta
=
\arg\min_{\pi\ge0}
\Big\{
\langle \bc_\theta,\pi\rangle
+
\varepsilon\,\mathrm{KL}\big(\pi\,\|\,\mu\!\otimes\!\nu\big)
+
\tau_1\,\mathrm{KL}\!\big(\pi\mathbf{1}\,\|\,\mu\big)
+
\tau_2\,\mathrm{KL}\big(\mathbf{1}^\top\pi\,\|\,\nu\big)
\Big\},
\]
where $\mu\in\mathbb{R}^K$ and $\nu\in\mathbb{R}^M$ are reference marginals. The relaxed column marginal permits multiple prototypes to concentrate on the same concept, naturally modeling many-to-one alignment. The relaxed row marginal allows surplus visual mass to shrink at finite cost, so unmatched background regions need not be forced into incorrect matches.

\noindent\textbf{Foreground-aware geometric suppression.} Although UOT permits mass variation, background-dominant prototypes can still introduce weak, noisy associations. We use CLS attention as a simple foreground prior: prototypes with low attention are treated as background. Let $\mathcal{B}\subseteq\{1,\dots,K\}$ denote the resulting index set. We then increase the cost of transporting background prototypes by
$\bc'_{kj}
=
\bc_{kj}
+
\lambda_{\mathrm{bg}}\,\mathbbm{1}(k\in\mathcal{B})$, where $\lambda_{\mathrm{bg}}$ controls suppression strength. This makes background transport uniformly more expensive, and under the unbalanced formulation the optimizer prefers to shrink background mass rather than match it to arbitrary concepts. Using $\bc'$, we compute the final plan $\pi_{\mathrm{VLOT}}\in\mathbb{R}^{K\times M}$, which defines the geometric coupling between visual prototypes and textual concepts.

The resulting coupling simultaneously (i) aggregates spatially coherent visual evidence, (ii) allows many-to-one semantic assignment, and (iii) suppresses unmatched background mass. This forms a geometry-aware cross-modal alignment tailored to vision–language concept modeling.
\black

\subsection{Learning Cross-Modal Geometry via Inverse OT}
\label{subsec:iot}

The geometric coupling defined in the previous section depends critically on the choice of cross–modal cost function $\bc_\theta(x,c)$. This function determines how visual evidence is measured against textual concepts and therefore shapes the induced transport plan. A central challenge is that visual and textual embeddings are produced by heterogeneous encoders and inhabit spaces with different geometric structures. Applying a fixed metric, such as cosine similarity or squared Euclidean distance, implicitly assumes that both modalities share a compatible embedding geometry. In practice, this assumption rarely holds, and mis-specified distances lead to distorted transport plans even under unbalanced formulations. 

\begin{figure}
\centering
\includegraphics[width=0.85\linewidth]{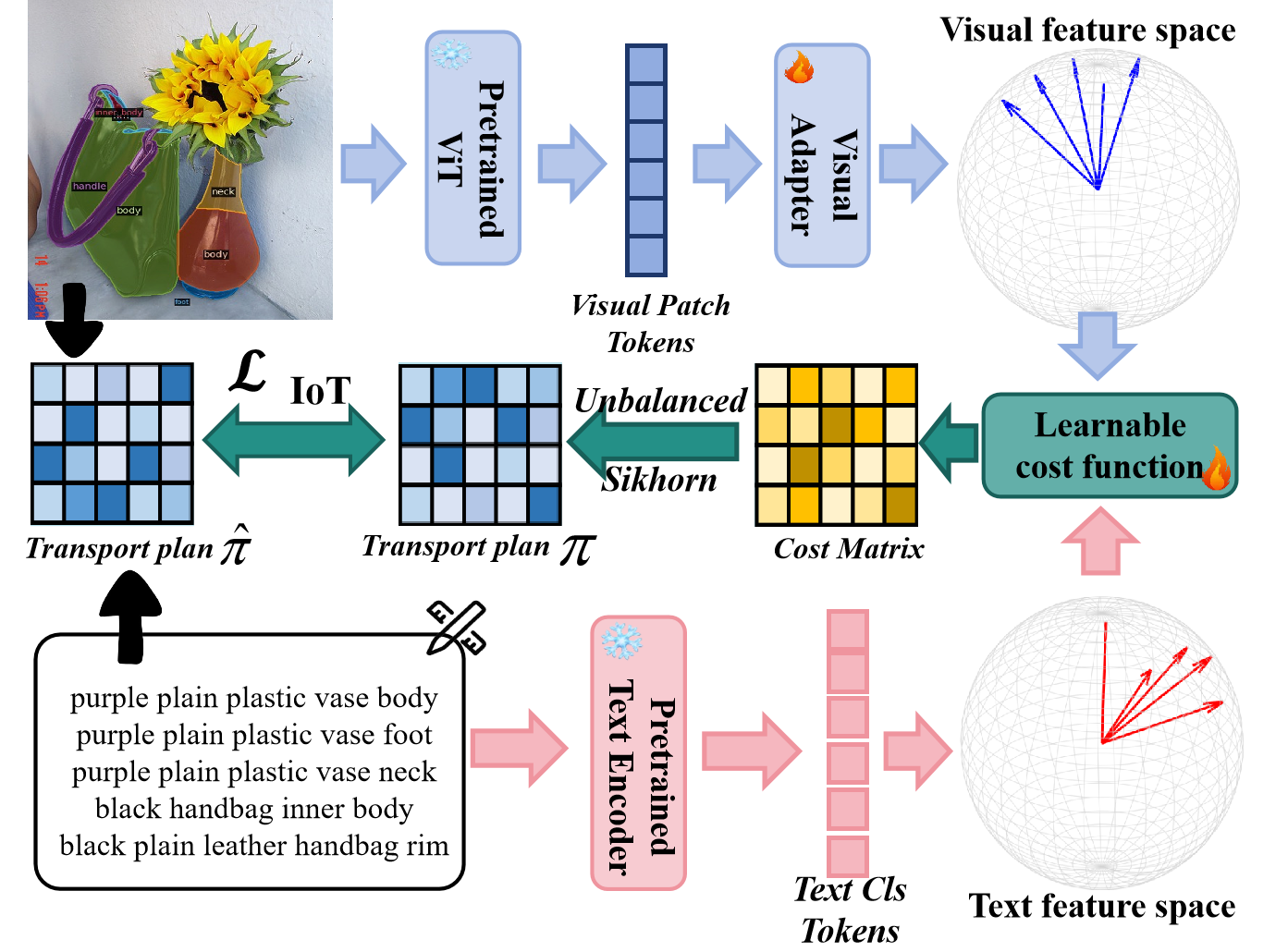}
\caption{\textbf{Training IoT cost functional.}
With object–component annotations datasets, we build ground-truth transport plans. A learnable multi-basis cost $\bc_\theta$ produces cost matrices, and unbalanced Sinkhorn plans are fitted to these labels to reflect true cross-modal distances between visual patches and text embeddings.}
\label{fig:2}
\vspace{-10pt}
\end{figure}

To this end, we learn the cross–modal geometry from data instead of prescribing it. Specifically, we adopt Inverse Optimal Transport (IoT) to infer a cost function whose induced transport plan agrees with empirical semantic correspondences. Unlike classical optimal transport, which fixes a cost and solves for an optimal plan, IoT reverses the direction: given observed couplings, it learns the cost under which those couplings become optimal.

For a given image, let $x_{1:N}\subset\mathcal{X}$ denote visual patches and $c_{1:M}\subset\mathcal{C}$ denote textual concepts. Suppose we have an empirical coupling $\hat\pi\in\mathbb{R}^{N\times M}$ derived from supervision. Given a parameterized cost matrix $(\bc_\theta)_{ij}=\bc_\theta(x_i,c_j)$, we compute the predicted plan $\pi_\theta$ by solving the unbalanced OT problem defined in Sec.~\ref{sec:vlot}. IoT then learns $\bc_\theta$ so that $\pi_\theta$ approximates $\hat\pi$, thereby inducing a geometry consistent with observed cross–modal structure.

In implementation, visual features $x\in\mathbb{R}^{d_v}$ and textual features $c\in\mathbb{R}^{d_t}$ are first mapped into a shared $d_p$-dimensional space. A learnable visual adapter $f_\psi$ projects visual tokens. The cost function is expressed as a linear combination of $K$ kernel bases $\Phi=\{\phi_k\}_{k=1}^K$ so that
\[
\bc_\theta(x,c)
=
\big\langle \boldsymbol{\theta},\,
\Phi\!\big(f_\psi(x),\,c\big)
\big\rangle,
\]
where $\boldsymbol{\theta}\in\mathbb{R}^K$ are learnable coefficients. The basis set includes squared Euclidean distance, Euclidean distance, cosine similarity, $1-\cos$, dot product, magnitude difference, and multi-scale RBF kernels (see Supplementary Material). This parameterization allows the model to represent a rich family of cross–modal geometries while retaining interpretability through weighted primitive distances.

To learn the cost, we minimize the discrepancy between the predicted plan $\pi_\theta$ and the empirical coupling $\hat\pi$. Direct Fenchel–Young alignment $\langle \bc_\theta,\pi_\theta-\hat\pi\rangle$ can be unstable when correspondences are sparse or noisy. We therefore adopt an absolute-weighted formulation:
$$
\mathcal{L}_{\mathrm{IoT}}
=
\big\|(\pi_\theta-\hat\pi)\odot \bc_\theta\big\|_{1,1}
+
\lambda_1\|\boldsymbol{\theta}\|_1,
$$
where $\odot$ denotes elementwise multiplication and $\|\cdot\|_{1,1}$ is the entrywise $\ell_1$ norm. When annotation masks $M$ are available, we compute $\|M\odot(\pi_\theta-\hat\pi)\odot \bc_\theta\|_{1,1}$. In practice, optimization is stabilized by detaching $\pi_\theta$ during early epochs before enabling full gradient propagation through the UOT solver. The learned cost $\bc_{\theta^\ast}$ (together with $f_{\psi^\ast}$) defines a frozen cross–modal geometry. This geometry is subsequently used by geometric coupling module and forward semantic flow stage to produce consistent and structure-aware concept alignment.

\subsection{From Discrete Coupling to Velocity-Based Concept Activation}\label{subsec:3.4}

The transport plan $\pi_{\mathrm{VLOT}}$ provides a discrete correspondence between visual prototypes and textual concepts under the learned cross–modal geometry. However, this coupling remains a static assignment matrix. While it identifies which regions align with which concepts, it does not describe how semantic evidence evolves within the embedding space, nor does it provide a continuous mechanism for measuring concept compatibility beyond the solved transport instance. In other words, $\pi_{\mathrm{VLOT}}$ captures \emph{where} mass moves, but not \emph{how} semantic structure is organized dynamically.

To obtain a geometry-consistent and generalizable representation, we lift discrete couplings into a continuous semantic flow. For prototype–concept pairs sampled according to $\pi_{\mathrm{VLOT}}$, define
\[
x_0=f_{\psi}(\tilde x_k)\in\mathbb{R}^{d_p},
\qquad
x_1=c_j\in\mathbb{R}^{d_p},
\]
and let $u=x_1-x_0$ denote the semantic displacement implied by optimal transport. Rather than treating alignment as a combinatorial mapping, we interpret these displacements as samples from an underlying continuous vector field.

We therefore learn a conditional velocity field $v_\phi(x,t,\mathrm{cond})$ that approximates these transport-induced directions in continuous time. For $t\sim\mathcal U[0,1]$, define
$x_t=(1-t)x_0+tx_1$ and  $u_t=x_1-x_0$, and  the velocity field is trained using flow matching, i.e.,
\begin{align}
\mathcal{L}_{\mathrm{FM}}
=
\mathbb{E}_{(x_0,x_1)\sim\pi_{\mathrm{VLOT}},\,t}
\big[
\|\,v_\phi(x_t,t,\mathrm{cond})-u_t\,\|^2
\big].
\end{align}
This objective promotes a smooth vector field whose local directions reproduce the semantic transport dynamics defined by the learned geometry.

In classical flow-based generative models, the learned velocity field is integrated via $\dot x=v_\phi(x_t,t)$ to recover the terminal state $x_1$, requiring numerical ODE solvers at inference. For Concept Bottleneck Models, such reconstruction is unnecessary. Our objective is to estimate concept activations, not to generate textual embeddings. From the theory of ordinary differential equations, under standard Lipschitz regularity, a velocity field uniquely determines trajectories; therefore, local velocity agreement already implies path consistency without explicit integration. Measuring instantaneous motion thus suffices to evaluate semantic alignment. $\pi_{\mathrm{VLOT}}$ is used only during training to define semantic displacements; inference depends solely on the learned velocity field and projected embeddings.

Further theoretical support arises from Schr\"odinger bridge formulations, the stochastic analogue of optimal transport flows. For Brownian motion conditioned on endpoints $X_0=x_0$ and $X_1=x_1$, the interpolating process admits the decomposition
$X_t=(1-t)x_0+t x_1 + B_t$,
where $B_t$ is a Brownian bridge with conditional variance $\sigma_t^2=t(1-t)$ \cite{chen2015stochastic,tong2023simulation}. The following lemma formalizes its midpoint property.

\begin{lemma}\label{lem:maximum}
Let $X_t$ be the Schr\"odinger bridge on $[0,1]$ for Brownian motion with endpoints $X_0=x_0$ and $X_1=x_1$. Then
$
X_t=(1-t)x_0+t x_1 + B_t,
$
where $B_t$ is a Brownian bridge with
$
\sigma^2_t(B_t)=t(1-t).
$
Thus, the variance is uniquely maximized at $t=\tfrac12$, so the midpoint is the most uncertain (and therefore most informative) stage for evaluating alignment.
\end{lemma}

The lemma implies that the midpoint concentrates maximal conditional uncertainty along the transport trajectory. Evaluating velocity consistency at this stage captures the richest semantic signal while regularizing trajectories toward smooth, geometry-consistent flows.

We therefore derive concept activations directly from instantaneous velocity agreement at $t=0.5$. For a projected visual prototype $x_0=f_{\psi^\ast}(\tilde x_k)$ and each concept embedding $x_{1,j}$ in the concept set, define the reference displacement
$u^{(k,j)}=x_{1,j}-x_0$ with x$_{1/2}^{(k,j)}=\tfrac12(x_0+x_{1,j})$.
The predicted midpoint velocity is
\[
v_{\mathrm{pred}}^{(k,j)}
=
v_\phi\!\big(x_{1/2}^{(k,j)},\,0.5,\,\mathrm{cond}\big).
\]
We measure alignment via
$S_{k,j}
=
-\big\|
v_{\mathrm{pred}}^{(k,j)}
-
u^{(k,j)}
\big\|^2$,
which quantifies how closely the predicted semantic motion matches the ideal displacement toward concept $j$ under the learned transport geometry. Unlike classical CBMs that rely on static feature similarity, this formulation evaluates \emph{velocity similarity}: activation increases when predicted motion aligns with the concept direction.

Concept activations are obtained by aggregating the strongest prototype responses 
$a_j
=
\frac{1}{K}
\sum_{k\in\mathrm{TopK}(S_{\cdot,j})}
S_{k,j}$.
This readout requires neither ODE integration nor solving an optimal transport problem at inference. Instead, it derives concept evidence from local motion consistency within the learned cross–modal geometry, providing a dynamic yet computationally efficient alternative to static similarity-based bottlenecks. Finally, the normalized activation vector is then passed through a linear classifier to predict class logits
\begin{equation}
\hat{y} = f_{\text{cls}}(\text{LayerNorm}(\mathbf{a})).
\end{equation}

\black

\section{Experimental results}\label{sec:experiments}
To assess the effectiveness of \textbf{OTF-CBM}, we evaluated from three perspectives:
\begin{enumerate}
\item[i)] Predictive accuracy and concept faithfulness on standard image classification benchmarks.
\item[ii)] Quality of cross–modal semantic flows, assessing how learned velocities transport visual evidence toward concept embeddings under the learned geometry.
\item[iii)] Robustness and object centricity under \emph{cross-modal semantic flows} shifts, using controlled background replacement and distribution changes to test that predictions rely on object evidence rather than context.
\end{enumerate}

\subsection{Datasets and Metrics}\label{subsec:setup}
\textbf{Evaluating CBM Performance.}
We evaluate on five image classification datasets spanning different granularities and domain complexity. \emph{CUB\mbox{-}200\mbox{-}2011} \cite{wah2011caltech} (11{,}788 bird images, 200 subcategories) and \emph{AwA2} \cite{xian2018zero} (37{,}322 animal images, 50 classes) are standard for interpretable concept learning. \emph{ImageNet\mbox{-}1K} \cite{deng2009imagenet} and \emph{CIFAR\mbox{-}100} \cite{krizhevsky2009learning} serve as large-scale benchmarks with diverse concepts and richer category structure. The scene-centric \emph{Places365} \cite{zhou2017places} probes robustness under contextual bias, since discriminative cues often lie in background rather than object appearance. Across all datasets, we report classification accuracy as the primary metric, assessing downstream performance and the effectiveness of the learned concept bottleneck.

\medskip
\noindent
\textbf{Evaluating Cross-Modal Semantic Flows.}
We report four metrics to assess the learned geometry and dynamics in our approach. Specifically, \emph{Transport Reconstruction Error} (TRE) measures how well the predicted transport plan recovers ground truth patch–concept correspondences, reflecting the fidelity of the learned cost geometry. \emph{Velocity Mean Squared Error} (VMSE) quantifies the discrepancy between predicted and theoretical semantic velocities along visual–to–textual trajectories. \emph{Mean Cosine Ratio} (MCR) evaluates directional consistency of the predicted flow field. \emph{Negative Pair Error} (NPE) captures the proportion of mismatched or reversed flow directions. Together, these metrics assess dynamic cross–modal flows beyond static embedding similarity. We present the definition of metrics for cross-modal transport and flow learning evaluation.
\begin{enumerate}
\item \textbf{TRE \cite{ma2020learning, li2019learning}}. Measures transport fidelity by comparing predicted and ground truth couplings~ by
$\displaystyle \mathrm{TRE}:=\frac{1}{B}\sum_{b=1}^{B}\big\|\pi_\theta^{(b)}-\hat\pi^{(b)}\big\|_{1}$.


\item \textbf{VMSE \cite{lipman2022flow}}. Measures the closeness between predicted instantaneous motion and ideal semantic displacement by
{\setlength{\abovedisplayskip}{2pt}%
 \setlength{\belowdisplayskip}{2pt}%
\begin{align}
\mathrm{VMSE}:=\frac{1}{N}\sum_{i=1}^{N}\Big\|\,v_\phi\!\big(x_{\tau_i}^{(i)},\tau_i\mid \cond^{(i)}\big)-u^{(i)}\Big\|_2^{2}.     
\end{align}}

\item \textbf{MCR \cite{liu2022flow}}. Quantifies directional consistency between predicted flow and target semantic direction.
{\setlength{\abovedisplayskip}{2pt}%
 \setlength{\belowdisplayskip}{2pt}%
\begin{align}
\mathrm{MCR}:=\frac{1}{N}\sum_{i=1}^{N}
\frac{\big\langle v_\phi\!\big(x_{\tau_i}^{(i)},\tau_i\mid \cond^{(i)}\big),\,u^{(i)}\big\rangle}
{\big\|v_\phi\!\big(x_{\tau_i}^{(i)},\tau_i\mid \cond^{(i)}\big)\big\|_2\,\|u^{(i)}\|_2}.     
\end{align}
}

\item \textbf{NPE \cite{tong2023improving}}. Reports the fraction of pairs with reversed alignment.
{\setlength{\abovedisplayskip}{2pt}%
 \setlength{\belowdisplayskip}{2pt}%
\begin{align}
\mathrm{NPE}:=\frac{1}{N}\sum_{i=1}^{N}
\mathbbm{1}\!\left[\ \big\langle v_\phi\!\big(x_{\tau_i}^{(i)},\tau_i\mid \cond^{(i)}\big),\,u^{(i)}\big\rangle<0\ \right].  
\end{align}
}
\end{enumerate}

\vspace{-0.2cm}
\subsection{Implementation Details}\label{subsec:implementation}
We use pre-trained \emph{DINOv2 ViT-L/14}~\cite{oquab2023dinov2} as the visual encoder $E_v$ with $d_v\!=\!1024$ and the pre-trained \emph{CLIP} text encoder $E_t$ with $d_t\!=\!768$~\cite{radford2021learning}. All images are resized to $224\times224$. A visual adapter $f_\psi$ (three-layer MLP with ReLU) projects features to the shared space with $d_p\!=\!768$. Optimization uses \emph{AdamW} with learning rate $1\times10^{-4}$ ($0.1$ decayeach epoch), weight decay $1\times10^{-5}$, batch size $64$, and $20$ training epochs. All experiments are implemented in PyTorch and run on two RTX~3090 GPUs. Additional hyperparameter details are in Supplementary Material.

\vspace{-0.3cm}
\subsection{Cost Learning Results}\label{subsec:cost-results}
\vspace{-0.1cm}
\textbf{Learned Cross-Modal Cost Functions.}
To characterize the inverse optimal transport (IoT) module, we examine the basis weights of the learned cost $\bc_\theta(x,c)$. The cost is a linear combination of 18 candidate bases, with the complete list given in the Supplementary Material. IoT experiment result reports the five largest contributors: squared angular distance, dot product similarity, and three hybrid terms (Dot–RBF, inverse quadratic, and root exponential). Together, these components capture local correlation and global semantic separation. The training objective is nonconvex, so different initializations can lead to local minimizer. In practice, runs that achieve similar validation loss induce nearly identical cross-modal geometries and alignments, indicating stable and interpretable learned metric.

\begin{table}
\vspace{-15pt}
\centering
\scriptsize
\renewcommand{\arraystretch}{1.3}
\setlength{\tabcolsep}{4pt}
\begin{tabular}{lcc}
\toprule
\textbf{Cost Function} & TRE $\downarrow$ & Remarks \\
\midrule
$\mathcal W^2_2$ & 2.371 & Standard metric \\
Cosine distance & 1.826 & Alignment baseline \\
\rowcolor{red!20}
IoT (ours) & 0.824 & Data-driven geometry \\
\bottomrule
\end{tabular}
\caption{TRE across different costs.}
\label{tab:tree}
\end{table}
\noindent\textbf{Comparison with Conventional Metrics.} We compare the learned IoT cost with two baselines: the Wasserstein distance and cosine similarity, both computed in the shared space. TRE measures how well the induced plan recovers annotated patch to concept couplings. As shown in Table~\ref{tab:tree}, IoT attains the lowest TRE and most closely matches the visual–textual correspondences, and the geometry strengthens downstream flow matching and concept inference.

\subsection{CBM + OT-FM Performance}\label{subsec:cbm-fm-results}

We compare our proposed \emph{OTF\mbox{-}CBM} with classical and recent concept bottleneck models (CBMs). Specifically, we include \emph{Vanilla\mbox{-}CBM}~\cite{koh2020concept}, which introduced the explicit concept layer and three training paradigms (independent, sequential, joint). Our two\mbox{-}stage optimization corresponds to the independent strategy reported as most interpretable in~\cite{koh2020concept}. We also evaluate \emph{CEM}~\cite{espinosa2022concept}, \emph{LaBo}~\cite{yang2023language}, \emph{SparseCBM}~\cite{semenov2024sparse}, \emph{CoopCBM}~\cite{sheth2023auxiliary}, and \emph{DOT\mbox{-}CBM}~\cite{xie2025discovering}. To eliminate confounding factors, all baselines are re\mbox{-}implemented with the same pre\mbox{-}trained ViT backbone and concept embeddings from the Label\mbox{-}Free CBM pipeline, ensuring consistent visual and conceptual representations. Results in Table~\ref{tab:cbm} show that our model attains the best performance across five benchmarks, with average top\mbox{-}1 accuracy gains of $+1.78\%$ (ImageNet), $+4.53\%$ (CUB), $+4.38\%$ (CIFAR\mbox{-}100), $+2.05\%$ (AwA2), and $+4.48\%$ (Places365).

\begin{table}[t]
\renewcommand{\arraystretch}{1.5}
\setlength{\tabcolsep}{8pt}
\centering
\resizebox{0.9\linewidth}{!}{
\begin{tabular}{lccccc}
\hline
\multicolumn{1}{c}{\multirow{2}{*}{\textbf{Method}}} 
& \multicolumn{5}{c}{\textbf{Classification Accuracy (↑)}} \\
\cline{2-6}
& ImageNet & CUB & CIFAR100 & AWA2 & Places365 \\
\hline
Vanilla-CBM & 79.17 & 78.32 & 80.04 & 93.15 & 44.80 \\
CEM         & 81.29 & 80.47 & 81.23 & 95.92 & 45.01 \\
LaBo        & 82.93 & 81.30 & 84.10 & 96.92 & 45.43 \\
SparseCBM   & 82.85 & 82.07 & 84.75 & 95.56 & 46.24 \\
CoopCBM     & 82.73 & 82.10 & 84.66 & 97.08 & 48.21 \\
DOT-CBM     & 83.84 & 85.39 & 85.83 & 96.83 & 50.65 \\
\hline
\rowcolor{red!20}
\textbf{OURS} & \textbf{85.62}\textcolor{ForestGreen}{(+1.78)} & \textbf{89.92}\textcolor{ForestGreen}{(+4.53)} & \textbf{90.21}\textcolor{ForestGreen}{(+4.38)} & \textbf{98.88}\textcolor{ForestGreen}{(+2.05)} & \textbf{55.13}\textcolor{ForestGreen}{(+4.48)} \\
\hline
\end{tabular}}
\caption{
\textbf{Classification performance across five datasets.}
}
\label{tab:cbm}
\end{table}

\noindent\textbf{Evaluating Cross-Modal Flow Modeling.}
We further assess the effectiveness of our cross\mbox{-}modal semantic flow components. Table~\ref{tab:flow-metrics} compares four configurations: \emph{Static OT Alignment} (no flow learning), \emph{OT Flow Matching} (linear flow), \emph{IoT + OT Flow Matching} (learned flow under balanced transport), and our full \emph{IoT + UOT Flow Matching} (learned many\mbox{-}to\mbox{-}one flow). Our method attains the lowest velocity mean squared error (VMSE~$\downarrow$), highest mean cosine ratio (MCR~$\uparrow$), and smallest negative pair error (NPE~$\downarrow$), indicating that the learned velocity field under the unbalanced transport geometry captures more faithful and directionally consistent cross\mbox{-}modal dynamics. The gains \emph{VMSE: 1.782\,$\to$\,1.000}, \emph{MCR: 0.884\,$\to$\,0.999}, and \emph{NPE: 0.197\,$\to$\,0.021} show that modeling many\mbox{-}to\mbox{-}one semantic flows enhances geometric fidelity and stabilizes the transport trajectory across modalities.
\begin{table}[t]
\centering
\renewcommand{\arraystretch}{1.45}
\setlength{\tabcolsep}{6pt}
\resizebox{0.8\linewidth}{!}{
\begin{tabular}{lcccc}
\hline
\textbf{Method} & \makecell{\textbf{Semantic Flow}\\\textbf{Modeling}} & \textbf{VMSE}~\textdownarrow & \textbf{MCR}~\textuparrow & \textbf{NPE}~\textdownarrow \\
\hline
Static OT Alignment & static matching & 2.467 & 0.739 & 0.434 \\
OT Flow Matching & Linear flow & 2.001 & 0.803 & 0.299 \\
\makecell[l]{IoT + \\OT Flow Matching} & Learned flow & 1.782 & 0.884 & 0.197 \\
\rowcolor{red!20}
\makecell[l]{\textbf{IoT +}\\\textbf{UOT Flow Matching (ours)}} &
\makecell[c]{\textbf{Learned flow}\\\textbf{(many-to-one)}} &
\textbf{1.000} & \textbf{0.999} & \textbf{0.021} \\
\hline
\end{tabular}}
\caption{
\textbf{Evaluation of semantic flow modeling.}
The proposed IoT+UOT formulation yields the most accurate and directionally consistent visual-to-concept flows.
}
\label{tab:flow-metrics}
\end{table}

\subsection{Ablations}\label{subsec:ablations}
Table~\ref{tab:ablation} reports ablation results on five benchmarks. Starting from a vanilla CBM baseline, replacing similarity-based  concept inference with classical OT consistently improves performance, indicating that explicit cross-modal coupling is beneficial. Our proposed Vision–Language OT (VLOT) further strengthens alignment by accounting for modality-specific structure. 

Learning the transport cost via inverse OT brings additional gains, highlighting the importance of accurately modeling cross-modal geometry. Finally, introducing the semantic flow mechanism yields the best overall performance, demonstrating that converting discrete transport plans into a continuous formulation improves robustness of concept inference. Overall, each component contributes cumulatively to the final results. Module hyperparameter choice ablations are provided in the Supplementary Material. We observe that the proposed framework is stable across a wide range of reasonable settings, with classification performance varying by less than $\pm1\%$ in normal regimes.

\begin{table}[t]
\centering
\renewcommand{\arraystretch}{1}
\setlength{\tabcolsep}{6pt}
\resizebox{0.95\linewidth}{!}{
\begin{tabular}{lccccc}
\hline
\multicolumn{1}{c}{}                         & \multicolumn{5}{c}{\textbf{Classification Accuracy}}                                               \\ \cline{2-6} 
\multicolumn{1}{c}{\multirow{-2}{*}{\textbf{Method}}} & ImageNet                     & CUB                          & CIFAR100 & AWA2  & PLACE365 \\ \hline
Vanilla-CBM                                  & 79.17                        & 78.32                        & 80.04    & 93.15 & 44.80    \\
with classical OT                & {\color[HTML]{333333} 80.42} & {\color[HTML]{333333} 80.31} & 85.23    & 94.38 & 47.7     \\
with Vision-Language OT          & 82.68                        & 84.31                        & 87.44    & 95.61 & 50.59    \\
\rowcolor{blue!20} +\textit{Inverse OT cost}             & {\color[HTML]{333333} 84.87} & {\color[HTML]{333333} 87.44} & 89.09    & 96.56 & 52.99    \\
\rowcolor{red!20} +\textit{Semantic Flow}               & 85.62                        & 89.92                        & 90.21    & 98.88 & 55.13    \\ \hline
\end{tabular}}
\caption{
\textbf{Ablation studies on key components.}
Each design improves classification performance, interpretability, or efficiency,
and their combination yields the best overall performance.}
\label{tab:ablation}
\end{table}

\begin{table}[t]
\centering
\renewcommand{\arraystretch}{1.35}
\setlength{\tabcolsep}{8pt}
\resizebox{0.8\linewidth}{!}{
\begin{tabular}{lcccc}
\toprule
\textbf{Method} & \multicolumn{2}{c}{\textbf{CUB}} & \multicolumn{2}{c}{\textbf{PLACES365}} \\
\cmidrule(lr){2-3}\cmidrule(lr){4-5}
 & \textbf{ID} & \textbf{OOD} & \textbf{ID} & \textbf{OOD} \\
\midrule
Vanilla-CBM & 78.3 & 30.1 & 44.8 & 21.3 \\
CEM         & 80.4 & 40.2 & 45.0 & 22.9 \\
LaBo        & 81.3 & 45.3 & 45.4 & 27.4 \\
SparseCBM   & 82.0 & 44.8 & 46.2 & 27.3 \\
CoopCBM     & 82.1 & 48.3 & 48.2 & 31.6 \\
DOT-CBM     & 85.3 & 67.5 & 50.6 & 42.1 \\
\midrule
\rowcolor{red!20}
\textbf{OURS} & \textbf{89.9}\textcolor{ForestGreen}{(+4.6)} & \textbf{82.0}\textcolor{ForestGreen}{(+14.5)} & \textbf{55.1}\textcolor{ForestGreen}{(+4.5)} & \textbf{50.5}\textcolor{ForestGreen}{(+8.4)} \\
\bottomrule
\end{tabular}}
\caption{
\textbf{Robustness evaluation under SAM-based background perturbations.}
Our method achieves the highest robustness across both in-domain (ID) and out-of-domain (OOD) conditions.
}
\label{tab:ood-results}
\end{table}

\subsection{Robustness and Generalization}\label{subsec:robustness}
Traditional CBMs often tie concepts to background or context, creating spurious correlations and weak OOD generalization. To probe this, we run a background-perturbation test: objects are segmented (e.g., birds on CUB), the background is replaced with randomly recolored noise, and the foreground is left intact. This creates an OOD shift that removes contextual bias while preserving semantic content; a concept-grounded model should retain accuracy.

Table~\ref{tab:ood-results} shows that OTF-CBM remains stable under this shift. On \emph{CUB}, OOD accuracy improves by $+14.5\%$ over the strongest baseline (DOT-CBM); on \emph{Places365}, the gain is $+8.4\%$. The ID–OOD gap shrinks substantially, indicating that cross-modal flow modeling anchors reasoning on object evidence rather than background shortcuts, yielding robust generalization to unseen environments.

\section{Conclusion}
We introduced OTF\mbox{-}CBM for cross-modal concept matching. A data-driven cost learned by inverse optimal transport corrects metric mismatch, and unbalanced OT handles many-to-one correspondences and background mass. On this geometry, a trained velocity field replaces static cosine similarity with midpoint velocity alignment, yielding efficient inference and spatially grounded activations. Experiments show consistent gains in classification accuracy, interpretability, and out-of-distribution generalization over prior CBMs. More importantly, our goal is not merely to improve classification accuracy in CBMs. Instead, we hope this work provides evidence that a potential performance bottleneck in vision-language models may stem from the joint optimization of heterogeneous modalities. Learning well-structured representations for each modality in its own embedding space before establishing accurate visual-text correspondences may offer a more effective solution to this bottleneck. We hope this perspective will inspire future research to further explore the combination of optimal transport and flow matching for more principled cross-modal representation learning.

\section*{Acknowledgement}
This work was supported by grants from the Natural Science Foundation of Shanxi Province (2024JCJCQN-66).

\bibliographystyle{splncs04}
\bibliography{references}

\end{document}